\documentclass{article}
\usepackage{theapa}
\usepackage{graphics,graphicx,amsmath}
\usepackage{doublespace}

\author{Anand Venkataraman\\
Computer Science, IIST\\Massey University\\
Palmerston North\\New Zealand\\mailto:A.Raman@massey.ac.nz}
\newcommand{\argmax}[1]{\underset{#1}{\rm argmax}\quad}

\newcommand{\W}{{\bf W}}
\newcommand{\Lex}{{\bf L}}
\newcommand{\A}{{\bf A}}
\newcommand{\p}{{\rm P}}
\newcommand{\C}{{\bf C}}
\newcommand{\X}{{\bf X}}

\begin{document}

\begin{singlespace}
\title{A Statistical Model for Word Discovery in Child Directed Speech}

\maketitle
\end{singlespace}

\begin{abstract}
\begin{quote}

A statistical model for segmentation and word discovery in child
directed speech is presented.  An incremental unsupervised learning
algorithm to infer word boundaries based on this model is described
and results of empirical tests showing that the algorithm is
competitive with other models that have been used for similar tasks
are also presented.

\end{quote}
\end{abstract}


\noindent {\bf Running head:\, }Word Discovery in Child Directed Speech

\newpage
\section{Introduction}
\label{sec:intro}

Speech lacks the acoustic analog of blank spaces that people are
accustomed to seeing between words in written text.  Young children
are thus faced with a non-trivial problem when trying to infer these
{\em word boundaries\/} in the often fluent speech that is directed at
them by adults, especially when they start out with no knowledge of
the inventory of words the language possesses.  Recent research has
shown statistical strategies to yield good results in inferring the
most likely word sequences from utterances.  This paper is concerned
with the development of one such statistical model for inferring word
boundaries in fluent child directed speech.  The model is formally
developed and described in Sections~\ref{sec:model}
and~\ref{sec:0freq}.  Section~\ref{sec:related} briefly discusses
related literature in the field and recent work on the same topic.
Section~\ref{sec:method} describes an unsupervised learning algorithm
based directly on the model developed in Section~\ref{sec:model}.
This section also describes the data corpus used to test the
algorithms and the methods used.  Results are presented and discussed
in Section~\ref{sec:results}.  Finally, the findings in this work are
summarized in Section~\ref{sec:summary}.

\section{Model Description}
\label{sec:model}

The segmentation model described in this section is an adaption from
\shortciteA{Jelinek:SMS97}.  Let $\A$ denote the acoustic evidence on
the basis of which the recognizer will make its decision about which
words were spoken.  In the present context, $\A$ is simply a
concatenation of symbols drawn from the set $\Sigma$ that constitutes
the phoneme inventory.  This is given in Appendix~A.  Every word $w_i$
is assumed to be made up of symbols drawn exclusively from this
phoneme inventory.

Let $\Lex$ denote a fixed and known vocabulary (hereafter
called the {\em lexicon}) of words, $w_i$, and
$$
\W = w_1, \cdots, w_n \qquad w_i \in \Lex
$$
denote a particular string of $n$ words belonging to the
lexicon $\Lex$.

If $\p(\W|\A)$ denotes the probability that the words $\W$ were spoken
given that the evidence $\A$ was observed, then the recognizer should
decide in favor of a word string $\hat{\W}$ satisfying
\begin{equation}
\label{eqn:w1}
\hat{\W} = \argmax{\W} \p(\W|\A)
\end{equation}
Using Bayes' rule, the right-hand side of Eqn~\ref{eqn:w1} can be
rewritten as
\begin{equation}
\p(\W|\A) = \frac{\p(\W) \p(\A|\W)}{\p(\A)}
\end{equation}
where $\p(\W)$ is the probability that word string $\W$ is uttered,
$\p(\A|\W)$ is the probability that the acoustic evidence $\A$ is
observed given that word string $\W$ was uttered and $\p(\A)$ is the
average probability that $\A$ will be observed.  It follows then that,
\begin{equation}
\hat{\W} = \argmax{\W} \p(\W) \p(\A|\W)
\end{equation}

Typically, the above quantity is computed using both acoustic and
language models.  $\p(\A|\W)$ is determined from the former and
$\p(\W)$ is determined from the latter.  However, acoustic models are
beyond the scope of this paper and so for the present purposes, we
assume that the acoustic evidence is a deterministic function of the
underlying word sequence, i.e. that there is only one way to
transcribe a given word phonemically.  Then $\p(\A|\W) = 1$.  So we
can write
\begin{equation}
\label{eqn:w2}
\hat{\W} = \argmax{\W} \p(\W)
\end{equation}
We can further develop Eqn~(\ref{eqn:w2}) by observing that
\begin{equation}
\p(\W) = \prod_{i=1}^n \p(w_i|w_1, \cdots, w_{i-1})
\end{equation}
where the right-hand side takes account of complete histories for each
word.  Assuming the existence of a many-to-one history function $\Phi$
that partitions the space of all possible histories into various
equivalence classes allows us to approximate at varying levels of
complexity.  We write
\begin{equation}
\p(\W) = \prod_{i=1}^n \p(w_i|\Phi(w_1, \cdots, w_{i-1}))
\end{equation}
or more simply just
\begin{equation}
\p(\W) = \prod_{i=1}^n \p(w_i|\Phi_{i-1})
\end{equation}
where we use $\Phi_{i-1}$ as shorthand for $\Phi(w_1, \cdots, w_{i-1})$.
If $C(\Phi)$ denotes the frequency of the particular history $\Phi$,
then we can make a first order estimate of the probability of a word
$w_i$ as follows.
\begin{equation}
\p(w_i|\Phi) = \frac{C(\Phi, w_i)}{C(\Phi)}
\end{equation}

We are now ready to introduce three practical approximations to the
required probability.  These are the unigram, bigram and trigram
models.  These classify histories as equivalent if they end in the
same $k$ words where $k$ is 0, 1 and 2 respectively.  The following
three equations give the probabilities making the unigram, bigram and
trigram assumptions.
$$
\begin{array}{ll}
\Phi_{i-1} = \langle \rangle & \Rightarrow
	\p(w_i|\Phi_{i-1}) = \p(w_i)\\
\Phi_{i-1} = \langle w_{i-1}\rangle & \Rightarrow 
	\p(w_i|\Phi_{i-1}) = \p(w_i|w_{i-1})\\
\Phi_{i-1} = \langle w_{i-2}, w_{i-1} \rangle & \Rightarrow 
	\p(w_i|\Phi_{i-1}) = \p(w_i|w_{i-2},w_{i-1})
\end{array}
$$
These yield three possibilities for testing
\begin{eqnarray}
\hat{\W} &=& \argmax{\W} \prod_{i=1}^n \p(w_i)\\
\hat{\W} &=& \argmax{\W} \p(w_1) \prod_{i=2}^n \p(w_i|w_{i-1})\\
\hat{\W} &=& \argmax{\W} \p(w_1) \p(w_2|w_1) \prod_{i=3}^n \p(w_i|w_{i-2},w_{i-1})
\end{eqnarray}
each of which we report results on.

In what follows, we estimate the required probabilities from tables of
unigrams, bigrams and trigrams.  Note that the unigram table is the
same as the lexicon $\Lex$.  Since the space of bigrams and trigrams
is considerably more sparsely populated than the space of unigrams, we
back-off to obtain probabilities of unseen bigrams and trigrams as
described in Section~\ref{sec:0freq}.

\section{Estimation of probabilities}
\label{sec:0freq}

We describe here the motivation and approach taken to address the
sparse data problem, namely that of estimating probabilities for
unseen words, bigrams and trigrams.  Suppose that probabilies are
estimated from relative frequencies.  Let $f(|)$ denote the relative
frequency function such that
$$
\begin{array}{lllll}
\p(w_i)          &=& f(w_i)          &\doteq& \frac{C(w_i)}
					{\sum\limits_{j=1}^{N}{C(w_j)}}\\
\p(w_i|w_{i-1})      &=& f(w_i|w_{i-1})     &\doteq&
				\frac{C(w_{i-1},w_i)}{C(w_{i-1})}\\
\p(w_i|w_{i-2}, w_{i-1}) &=& f(w_i|w_{i-2},w_{i-1}) &\doteq&
			\frac{C(w_{i-2},w_{i-1},w_i)}{C(w_{i-2},w_{i-1})}
\end{array}
$$
where $N$ is the number of distinct words in $\Lex$ and $C(w_i),
C(w_{i-1}, w_i)$ and $C(w_{i-2}, w_{i-1}, w_i)$ are the frequencies of
the unigram $w_i$, bigram $w_{i-1}, w_i$ and trigram $w_{i-2},
w_{i-1}, w_i$ in their respective tables.\footnote{We could use
separate relative frequency functions $f_\Sigma, f_1, f_2$ and $f_3$
and separate count functions $C_\Sigma, C_1, C_2$ and $C_3$ to
distinguish relative frequencies and counts from the phoneme, unigram,
bigram and trigram tables.  But in the following the context makes it
adequately clear which entity each function refers to and where the
counts are obtained from.  So we omit the subscripts in favor of
clarity.}  It now quickly becomes obvious that this method assigns
zero probability to any segmentation that contains an as yet unseen
word, bigram or trigram.  This becomes particularly problematic in an
incremental learning algorithm which starts out with no domain
knowledge whatsoever, i.e. the $n$-gram tables are initially empty.
They are only populated as a result of words that are inferred from
input utterances.  What we need then is a method that can reliably
assign reasonable non-zero probabilities to all events, novel or not.

The general problem of estimating probabilities for unseen events has
been studied in depth and is also the subject of much current
research.  Recently, for example, \citeA{Dagan:SBM99} uses a
similarity based model for estimating bigram probabilities.  In a
similar vein, we could estimate the probability of previously unseen
$n$-grams from probabilities of similar words in similar contexts.
However, it is difficult to do this in the relatively sparser
vocabulary of child-directed speech.  So we instead resort to simply
using a version of the Katz back-off scheme \cite{Katz:EPF87}.  This
is an adaptation of Method C suggested and evaluated by
\shortciteA{Witten:ZFP91}.  A portion of the probability space, which
we call {\em escape space}\/ is reserved for previously unseen
$n$-grams.  When a novel $n$-gram is encountered, its probability is
the proportion of this escape space determined by the probability of
the $n$-gram computed from some distribution over just the novel
$n$-grams.  In terms of adaptive text compression or coding theory,
one could imagine encoding a novel $n$-gram by first encoding an {\em
escape}\/ code and then encoding the novel $n$-gram by means of some
other prearranged alternate scheme.  The difference between various
schemes that utilize this technique is really in how much of the
original probability space is reserved for the novel $n$-gram,
i.e. what the probability of observing a novel $n$-gram is.

Consider unigrams first.  We define the probability of a novel word to
be the product of the probabilities of its individual phonemes in
sequence followed by a sentinel.  The sentinel phoneme, denoted by
``\#'' $\not\in \Sigma$, is introduced in order to normalize the
distribution over the space of all possible unigrams.  The phoneme
inventory is fixed and known beforehand (See Appendix~A). So the
zero-frequency problem cannot recur here.

The exact amount of escape space reserved for novel words in Method C
of \shortciteA{Witten:ZFP91} varies dynamically in an algorithm that
employs this technique.  In particular, novelty is seen as an event in
its own right.  Thus it is assigned a probability of $r/(n+r)$ where
$r$ is the total number of times a novel word has been seen in the
past, which is just the size of the lexicon (each word in the lexicon
must have been a novel word when it was first introduced) and $n$ is
the sum of the frequencies of all the words in $\Lex$.  Thus if $k$ is
the length in phonemes, excluding the sentinel, of an arbitrary word
$w$ and $w[j]$ is its $j$th phoneme, the probability of $w$ is given
by
\begin{equation}
\label{eqn:pnovel}
\p(w) = \frac{N f(\#) \prod\limits_{j=1}^{k}f(w[j])}
	{(1-f(\#))(N + \sum\limits_{i=1}^{N}C(w_i))}
\end{equation}
if $w$ is novel and 
\begin{equation}
\label{eqn:pfamiliar}
\p(w) = \frac{C(w)}{N + \sum\limits_{i=1}^{N}C(w_i)}
\end{equation}
if it is not.  The normalization by dividing using $1-f(\#)$
in~(\ref{eqn:pnovel}) is necessary because otherwise
\begin{eqnarray}
\sum_w \p(w) &=& \sum_{i=1}^\infty (1-\p(\#))^i\p(\#)\\
&=& 1-\p(\#)
\end{eqnarray}
Since we estimate $\p(w[j])$ by $f(w[j])$, dividing by $1-f(\#)$ will ensure
that $\sum_w \p(w) = 1.$

Bigrams and trigrams are handled similarly.  We give a more formal and
recursive statement of the estimation as follows.
\begin{eqnarray}
\p(w_i|w_{i-2},w_{i-1}) &=& \left\{ \begin{array}{ll}
\alpha_3 Q_3(w_i|w_{i-2},w_{i-1}) & {\rm if\ } C(w_{i-2},w_{i-1},w_i) >K \\
\beta_3 \p(w_i|w_{i-1}) & {\rm otherwise}\\
\end{array} \right. \label{eqn:smooth3}\\
\p(w_i|w_{i-1}) &=& \left\{ \begin{array}{ll}
\alpha_2 Q_2(w_i|w_{i-1}) & {\rm if\ } C(w_{i-1},w_i) >L \\
\beta_2 \p(w_i) & {\rm otherwise}\\
\end{array} \right. \label{eqn:smooth2}\\
\p(w_i) &=& \left\{ \begin{array}{ll}
\alpha_1 Q_1(w_i) & {\rm if\ } C(w_i) >M \\
\beta_1 \p_\Sigma(w_i) & {\rm otherwise}\\
\end{array} \right. \label{eqn:smooth1}
\end{eqnarray}
where the $Q_i$ are Good-Turing type functions, $\alpha_i$ and
$\beta_i$ are chosen so as to normalize the trigram, bigram and
unigram probability estimates.  Constants $K, L$ and $M$ are suitable
thresholds.  

Now let $\Phi$ be a classifier that partitions the space of trigrams
into two equivalence classes such that the trigram $w_1,w_2,w_3$
belongs to $\Phi_{\le K}$ if $C(w_1,w_2,w_3) \le K$ and to $\Phi_{>K}$
otherwise.  Then we can set $\alpha_3$ and $\beta_3$ to just the
occupancy probabilities of $\Phi_{\le K}$ and $\Phi_{>K}$ so that
$\p(x,\Phi) = \p(\Phi)\p(x|\Phi)$.  If we now let $K = 0$, then $\beta_3$
is just the probability that the observed trigram is novel, which we
may estimate by relative frequency as
\begin{eqnarray}
\beta_3 &=& \frac{N_3}{N_3 + \sum\limits_{w_1,w_2,w_3} C(w_1,w_2,w_3)}
\end{eqnarray}
where $N_3$ is the number of unique trigrams.\footnote{We mean the
probability of the trigram itself, not the probability of the third
word given the first two.} Likewise, letting $L=M=0$, we obtain
\begin{eqnarray}
\beta_2 &=& \frac{N_2}{N_2 + \sum\limits_{w_1,w_2} C(w_1,w_2)}\\
\beta_1 &=& \frac{N_1}{N_1 + \sum\limits_{w_1} C(w_1)}
\end{eqnarray}
with $\alpha_i = 1-\beta_i$ in general.  Note now that
formulas~(\ref{eqn:pnovel}) and~(\ref{eqn:pfamiliar}) follow directly
from Eqn~(\ref{eqn:smooth1}) when we let $Q_1$ be the relative
frequency estimator for previously observed unigrams, and $N_1 = N$ is
the size of the lexicon.  For example, 
$$
\alpha_1 = \frac{\sum_{w_1} C(w_1)}{N_1 + \sum_{w_1} C(w_1)},
	\quad Q_1(w_1) = \frac{C(w_1)}{\sum_{w_1} C(w_1)}
$$ 
giving
$$
\p(w_1) = \alpha_1Q_1(w_1) = \frac{C(w_1)}{N_1 + \sum_{w_1}C(w_1)}
$$ 
which is precisely the quantity in~(\ref{eqn:pfamiliar}) up to
renaming.  Below we summarize the formulas for calculating unigram,
bigram and trigram probabilities.  These have been obtained as
described in the preceding discussion.

\begin{eqnarray}
\p(w_i|w_{i-2},w_{i-1}) &=& \left\{
	\begin{array}{ll}
	\frac{S_3}{N_3 + S_3} \frac{C(w_{i-2},w_{i-1},w_i)}{C(w_{i-1},w_i)} &
	{\rm if\ } C(w_{i-2},w_{i-1},w_i) > 0 \\
	\frac{N_3}{N_3 + S_3}\p(w_i|w_{i-1}) & {\rm otherwise}\\
	\end{array}
	\right. \label{eqn:smooth3-final}\\
\p(w_i|w_{i-1}) &=& \left\{
	\begin{array}{ll}
	\frac{S_2}{N_2 + S_2} \frac{C(w_{i-1},w_i)}{C(w_i)} &
	{\rm if\ } C(w_{i-1},w_i) > 0 \\
	\frac{N_2}{N_2 + S_2} \p(w_i) & {\rm otherwise}\\
	\end{array}
	\right. \label{eqn:smooth2-final}\\
\p(w_i) &=& \left\{
	\begin{array}{ll}
	\frac{C(w_i)}{N_1 + S_1} & {\rm if\ } C(w_i) > 0 \\
	\frac{N_1}{N_1 + S_1} \p_\Sigma(w_i) & {\rm otherwise}\\
	\end{array}
	\right. \label{eqn:smooth1-final}\\
\p_\Sigma(w_i) &=& \frac{f(\#)\prod\limits_{j=1}^{k} f(w_i[j])}{1-f(\#)}
\end{eqnarray}
where as before, $N_3, N_2$ and $N_1$ denote the number of unique
previously observed trigrams, bigrams and unigrams respectively, $S_3
= \sum_{w_1,w_2,w_3} C(w_1,w_2,w_3)$ is the sum of the frequencies of
all observed trigrams, $S_2 = \sum_{w_1,w_2} C(w_1,w_2)$ is the sum of
the frequencies of all observed bigrams and $S_1 = \sum_{w_1} C(w_1)$
is the sum of the frequencies of all observed unigrams.  $k$ denotes
the length of word $w_i$, excluding the sentinel character, `\#', and
$w_i[j]$ denotes its $j$th phoneme.

\section{Related work}
\label{sec:related}

Model Based Dynamic Programming, hereafter referred to as MBDP-1
\cite{Brent:EPS99}, is probably the most recent work that addresses
the exact same issue as that considered in this paper.  Both the
approach presented in this paper and Brent's MBDP-1 are based on
explicit probability models.  Approaches not based on explicit
probability models include those based on information theoretic
criteria such as MDL \shortcite{Brent:DRP96,deMarcken:UAL95},
transitional probability \shortcite{Saffran:WSR96} or simple recurrent
networks \shortcite{Elman:FST90,Christiansen:LSS98}.  The maximum
likelihood approach due to \citeA{Olivier:SGL68} is probabilistic in
the sense that it is geared towards explicitly calculating the most
probable segmentation of each block of input utterances.  However, it
is not based on a formal statistical model.  To avoid needless
repetition, we only describe Brent's MBDP-1 below and direct the
interested reader at \citeA{Brent:EPS99} which provides an excellent
review of many of the algorithms mentioned above.

\subsection{Brent's model based dynamic programming method}

\citeA{Brent:EPS99} describes a model based approach to inferring word
boundaries in child-directed speech.  As the name implies this
technique uses dynamic programming to infer the best segmentation.  It
is assumed that the entire input corpus consisting of a concatenation
of all utterances in sequence is a single event in probability space
and that the best segmentation of each utterance is implied by the
best segmentation of the corpus itself.  The model thus focuses on
explicitly calculating probabilities for every possible segmentation
of the entire corpus, subsequently picking that segmentation with the
maximum probability.  More precisely, the model attempts to calculate
$$
\p(\bar{w}_m) = \sum_n \sum_L \sum_f \sum_s
	\p(\bar{w}_m|n,L,f,s) \cdot \p(n,L,f,s)
$$
for each possible segmentation of the input corpus where the left
hand side is the exact probability of that particular segmentation of
the corpus into words $\bar{w}_m = w_1 w_2 \cdots w_m$ and the sums
are over all possible numbers of words, $n$, in the lexicon, all
possible lexicons, $L$, all possible frequencies, $f$, of the
individual words in this lexicon and all possible orders of words,
$s$, in the segmentation.  In practice, the implementation uses an
incremental approach which computes the best segmentation of the
entire corpus upto step $i$, where the $i$th step is the corpus upto
and including the $i$th utterance.  Incremental performance is thus
obtained by computing this quantity anew after each segmentation $i-1$,
assuming, however, that segmentations of utterances upto but not
including $i$ are fixed.

There are two problems with this approach.  Firstly, the assumption
that the entire corpus of observed speech be treated as a single event
in probability space appears both radical and unsubtantiated in
developmental studies.  Indeed, it seems reasonable to suppose that a
child will use the entire arsenal of resources at its disposal to try
and make sense of each individual utterance directed at it and
immediately make available any knowledge gleaned from this process for
the next segmentation task.  This fact is appreciated even in
\citeA[p.89]{Brent:EPS99} which states ``{\em From a cognitive
perspective, we know that humans segment each utterance they hear
without waiting until the corpus of all utterances they will ever hear
becomes available}.''  Thus although the incremental algorithm in
\citeA{Brent:EPS99} is consistent with a developmental model, the
formal statistical model of segmentation is not.  Secondly, making
this assumption increases the computational complexity of the
incremental algorithm significantly.  The approach presented in this
paper circumvents these problems through the use of a conservative
statistical model that is directly implementable as an incremental
algorithm.  In the following section, we describe how the model and
its 2-gram and 3-gram extensions are adapted for implementation and
describe the experimental and scoring setups.

\section{Method}
\label{sec:method}

As in \shortciteA{Brent:EPS99}, the model developed in
Sections~\ref{sec:model} and~\ref{sec:0freq} is presented as an
incremental learner.  The only knowledge built into the system at
start-up is the phoneme table with a uniform distribution over all
phonemes, including the sentinel phoneme.  The learning algorithm
considers each utterance in turn and computes the most probable
segmentation of the utterance using a Viterbi search
\shortcite{Viterbi:EBC67} implemented as a dynamic programming
algorithm described shortly.  The most likely placement of word
boundaries computed thus is committed to before considering the next
presented utterance.  Committing to a segmentation involves learning
unigram, bigram and trigram, as well as phoneme frequencies from
the inferred words.  These are used to update the respective tables.

To account for effects that any specific ordering of input utterances
may have on the segmentations output, the performance of the algorithm
is averaged over 1000 runs, with each run being input a random
permutation of the input corpus.  Since this is not done in
\shortciteA{Brent:EPS99}, unaveraged results from a single run are
also presented for purposes of comparison.

\subsection{The Input Corpus}

The corpus, which is identical to the one used by
\shortciteA{Brent:EPS99}, consists of orthographic transcripts made by
\citeA{Bernstein:PPC87} from the CHILDES collection
\cite{MACWHINNEY:CLD85}.  The speakers in this study were nine mothers
speaking freely to their children, whose ages averaged 18 months
(range 13--21).  Brent and his colleagues also transcribed the corpus
phonemically (using the ASCII phonemic representation in Appendix~A)
ensuring that the number of subjective judgments in the pronunciation
of words was minimised by transcribing every occurrence of the same
word identically.  For example, ``look,'' ``drink'' and ``doggie''
were always transcribed ``lUk,'' ``drINk'' and ``dOgi'' regardless of
where in the utterance they occurred and which mother uttered them in
what way.  The corpus, thus transcribed, consists of a total of 9790
such utterances and 33397 characters including one space between each
pair of words and one newline after each utterance.  For purposes of
illustration, Table~\ref{tbl:corpus} lists the first 20 such
utterances from a random permutation of this corpus.

\begin{table}
\begin{center}
\begin{tabular}{|l|l|} \hline
{\bf Phonemic Transcription} & {\bf Orthographic English text} \\ \hline \hline
hQ sIli 6v mi & How silly of me \\
lUk D*z 6 b7 wIT hIz h\&t & Look, there's the boy with his hat \\
9 TINk 9 si 6nADR bUk & I think I see another book \\
tu & Two \\
DIs wAn & This one \\
r9t WEn De wOk & Right when they walk \\
huz an D6 tEl6fon \&lIs & Who's on the telephone, Alice? \\
sIt dQn & Sit down \\
k\&n yu fid It tu D6 dOgi & Can you feed it to the doggie? \\
D* & There \\
du yu si hIm h( & Do you see him here? \\
lUk & Look \\
yu want It In & You want it in \\
W* dId It go & Where did it go? \\
\&nd WAt \# Doz & And what are those? \\
h9 m6ri & Hi Mary \\
oke Its 6 cIk & Okay it's a chick \\
y\& lUk WAt yu dId & Yeah, look what you did \\
oke & Okay \\
tek It Qt & Take it out \\ \hline
\end{tabular}
\end{center}
\caption{Twenty randomly chosen utterances from the input corpus with
their orthographic transcripts.  See Appendix~A for a list of the
ASCII representations of the phonemes.}
\label{tbl:corpus}
\end{table}

\subsection{Algorithm}

The dynamic progamming algorithm finds the most probable word sequence
for each input utterance by assigning to each utterance a score equal
to its probability and committing to the utterance with the highest
score.  In practice, the implementation computes the negative log of
this score and thus commits to the utterance with the least negative
log of the probability.  The algorithm for the unigram language model
is presented in recursive form in Figure~\ref{fig:dyn-rec}.  An
iterative version, which is the one actually implemented, is also
shown in Figure~\ref{fig:dyn-it}.  Algorithms for bigram and trigram
language models are straightforward extensions of that given for the
unigram model.

\begin{figure}[htb]
\begin{small}
\subsubsection{Algorithm: evalUtterance}
\begin{verbatim}
BEGIN
   Input (by ref) utterance u[0..n] where u[i] are the characters in it.

   bestSegpoint := n;
   bestScore := evalWord(u[0..n]);
   for i from 0 to n-1; do
      subUtterance := copy(u[0..i]);
      word := copy(u[i+1..n]);
      score := evalUtterance(subUtterance) + evalWord(word);
      if (score < bestScore); then
         bestScore = score;
         bestSegpoint := i;
      fi
   done
   insertWordBoundary(u, bestSegpoint)
   return bestScore;
END
\end{verbatim}
\end{small}
\caption{Recursive optimisation algorithm to find the best
segmentation of an input utterance using the unigram language model
described in this paper.}
\label{fig:dyn-rec}
\end{figure}

\begin{figure}[htb]
\begin{small}
\subsubsection{Function: evalWord}
\begin{verbatim}
BEGIN
   Input (by reference) word w[0..k] where w[i] are the phonemes in it.

   score = 0;
   if L.frequency(word) == 0; then {
      escape = L.size()/(L.size()+L.sumFrequencies())
      P_0 = phonemes.relativeFrequency('#');
      score = -log(escape) -log(P0/(1-P0));
      for each w[i]; do
         score -= log(phonemes.relativeFrequency(w[i]));
      done
   } else {
      P_w = L.frequency(w)/(L.size()+L.sumFrequencies());
      score = -log(P_w);
   }
   return score;
END
\end{verbatim}
\end{small}
\caption{The function to compute $-\log \p(w)$ of an input word $w$.
L stands for the lexicon object.  If the word is novel, then it
backs off to a using a distribution over the phonemes in the word.}
\label{fig:word-len}
\end{figure}

\begin{figure}[htb]
\begin{small}
\subsubsection{Algorithm: evalUtterance}
\begin{verbatim}
BEGIN
   Input (by ref) utterance u[0..n] where u[i] are the phonemes in it.
   Array evalUtterance[0..n];
   Array previousBoundary[0..n];

   for i from 0 to n-1; do
      evalUtterance[i] := evalWord(u[0..i]);
      prevBoundary[i] := -1;
      for j from 0 to i; do
         score := evalUtterance[j] + evalWord(u[j+1..i]);
         if (score < evalUtterance[i]); then
            evalUtterance[i] := score;
            prevBoundary[i] := j;
         fi
      done
   done
   i = n-1;
   while i >= 0; do
     insertWordBoundary(u,prevBoundary[i]);
     i := prevBoundary[i];
   done
   return evalUtterance[n];
END
\end{verbatim}
\end{small}
\caption{Iterative version of the Algorithm in Figure~\ref{fig:dyn-rec}.}
\label{fig:dyn-it}
\end{figure}

One can easily see that the running time of the algorithm is $O(mn^2)$
in the total number of utterances ($m$) and the length of each
utterance ($n$) assuming an efficient implementation of a hash table
allowing nearly constant lookup time is available.  Since individual
utterances typically tend to be small, especially in child-directed
speech as evidenced in Table~\ref{tbl:corpus}, the algorithm
practically approximates to a linear time procedure.  A single run
over the entire corpus typically completes in under 10 seconds on an
i686 based PC running Linux 2.2.5-15.

Although the algorithm is presented as an unsupervised learner, a
further experiment to test the responsiveness of each algorithm to
training data is also reported on.  The procedure involved reserving
for training increasing amounts of the input corpus from 0\% in steps
of approximately 1\% (100 utterances).  During the training period,
the algorithm is presented the correct segmentation of the input
utterance which it uses to update trigram, bigram, unigram and phoneme
frequencies as required.  After the initial training segment of the
input corpus has been considered, subsequent utterances are then
processed in the normal way.

\subsection{Scoring}
\label{sec:scoring}

In line with the results reported in \shortciteA{Brent:EPS99}, three
scores are reported --- precision, recall and lexicon precision.
Precision is defined as the proportion of predicted words that are
actually correct.  Recall is defined to be the proportion of correct
words that were predicted.  Lexicon precision is defined to be the
proportion of words in the predicted lexicon that are correct.  In
addition to these, the number of correct and incorrect words in the
predicted lexicon were also computed, but they are not graphed here
because the lexicon precision is a good indicator of both.

Precision and recall scores were computed incrementally and
cumulatively within scoring blocks each of which consisted of 500
consecutive utterances in the non-averaged case and 100 utterances in
the averaged case.  These scores are computed only for the utterances
within each block scored and thus they represent the performance of
the algorithm only on the block scored, occurring in the exact context
among the other scoring blocks.  Lexicon scores carried over blocks
cumulatively.  Precision, recall and lexicon scores of the algorithm
in the case when it used various amounts of training data are computed
over the entire corpus.  All scores are reported as percentages.

\section{Results}
\label{sec:results}

\begin{figure}[htb]
\begin{center}
  \includegraphics[width=11.8cm]{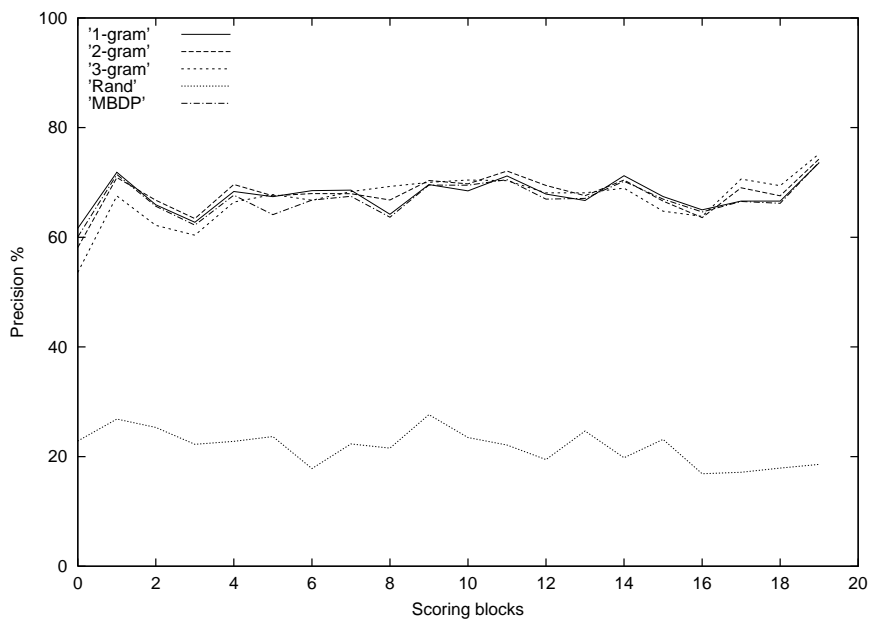}
  \caption{Precision, given by the percentage of identified words that
  are correct, as measured against the target data.  The horizontal
  axis represents the number of blocks of data scored, where each
  block represents 500 utterances. The plots show the performance of
  the 1-gram, 2-gram, 3-gram and MBDP-1 algorithms and also a random
  baseline which is given the correct number of word boundaries.}
  \label{fig:pre}
\end{center}
\end{figure}

\begin{figure}[htb]
\begin{center}
  \includegraphics[width=11.8cm]{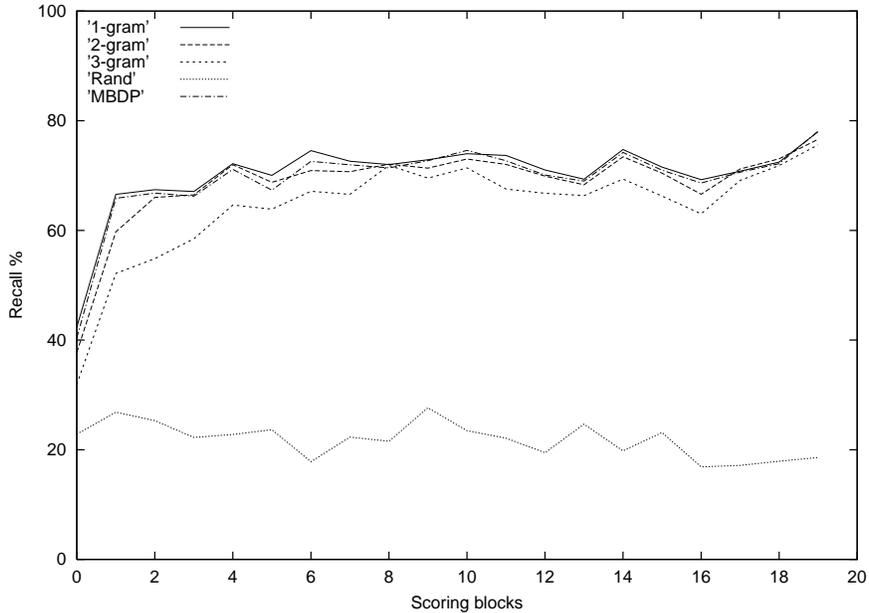}
  \caption{Recall, given by the percentage of words in the
  target data that were identified correctly by the algorithm.}
  \label{fig:rec}
\end{center}
\end{figure}

\begin{figure}[htb]
\begin{center}
  \includegraphics[width=11.8cm]{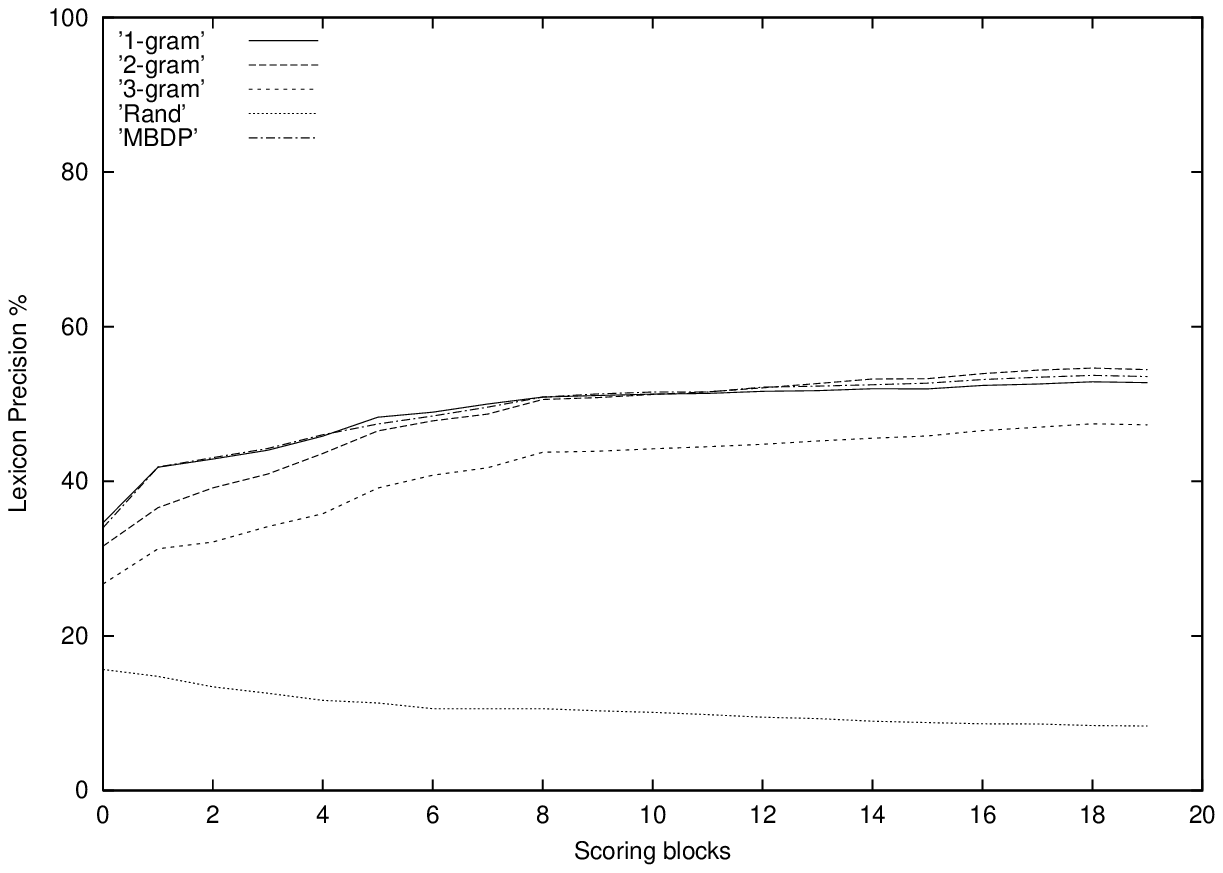}
  \caption{Lexicon precision, defined as the percentage of words in 
  the learned lexicon that are also in the target lexicon at that point.}
  \label{fig:lex}
\end{center}
\end{figure}

Figures~\ref{fig:pre}--\ref{fig:lex} plot the precision, recall and
lexicon precision of the proposed algorithm for each of the unigram,
bigram and trigram models against both the MBDP-1 algorithm and the
same random baseline as in \shortciteA{Brent:EPS99}.  This baseline
algorithm is given an important advantage --- It knows the exact
number of word boundaries, although it doesn't know their locations.
Brent argued that if MBDP-1 performs as well as this random baseline,
then at the very least, it suggests that the algorithm is able to
infer the right number of word boundaries.  The results presented here
can be directly compared with the performance of related algorithms
due to \shortciteA{Elman:FST90} and \shortciteA{Olivier:SGL68} because
\shortciteA{Brent:EPS99} reports results on them over exactly the same
corpus.

\subsection{Smoothing of Results}

\begin{figure}[htb]
\begin{center}
  \includegraphics[width=11.8cm]{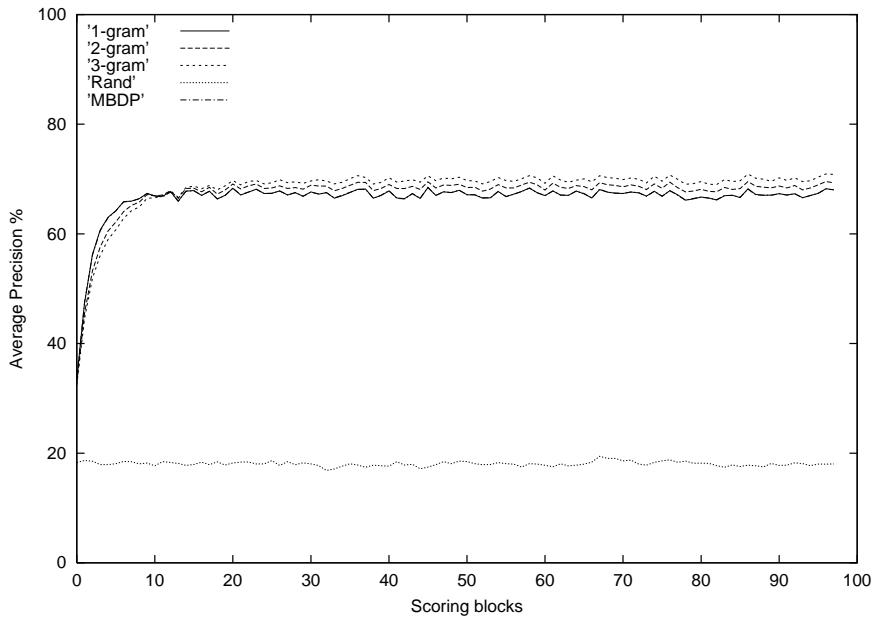}
  \caption{Averaged precision.  This is a plot of the segmentation
  precision over 100 utterance blocks averaged over 1000 runs
  each using a random permutation of the input corpus.}
  \label{fig:r-pre}
\end{center}
\end{figure}

\begin{figure}[htb]
\begin{center}
  \includegraphics[width=11.8cm]{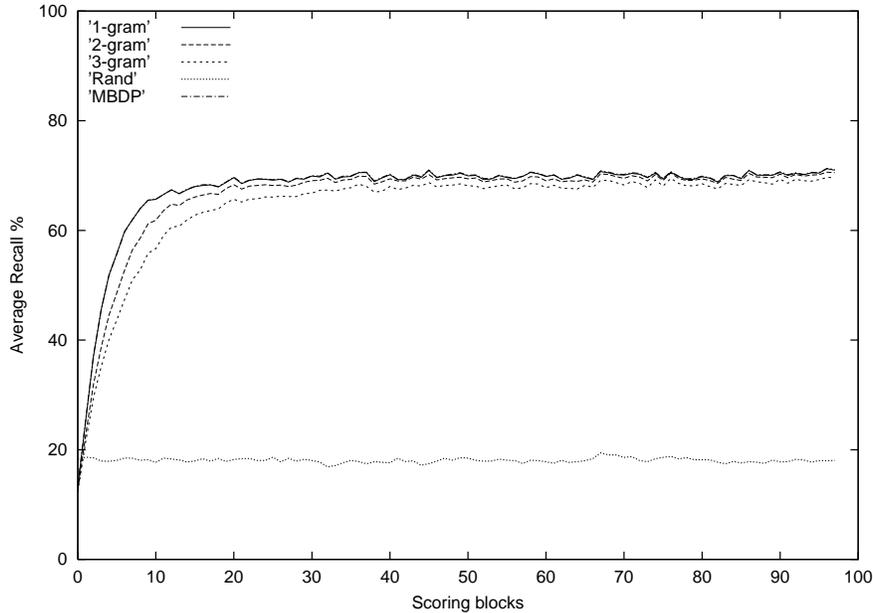}
  \caption{Averaged Recall over 1000 runs each using a random
  permutation of the input corpus.}
  \label{fig:r-rec}
\end{center}
\end{figure}

\begin{figure}[htb]
\begin{center}
  \includegraphics[width=11.8cm]{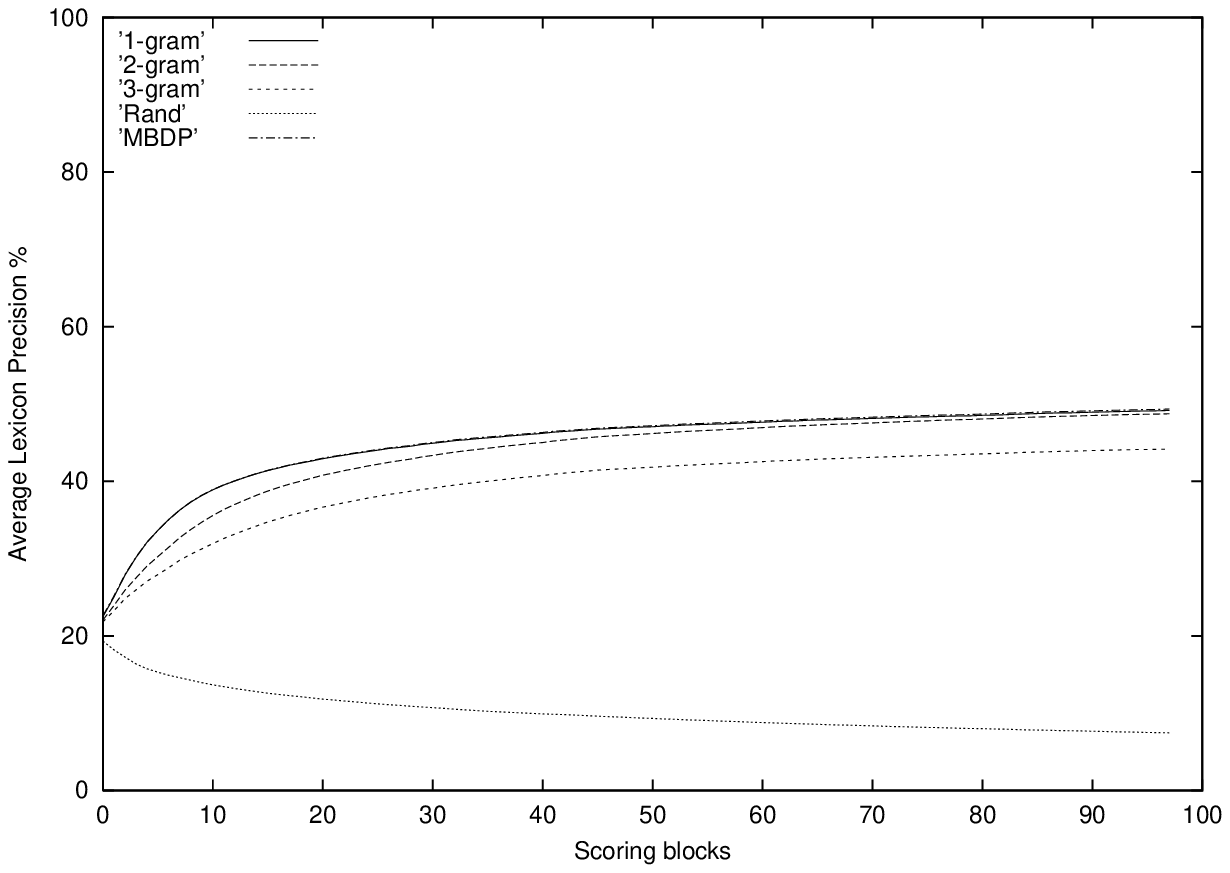}
  \caption{Averaged lexicon precision over 1000 runs each using a
  random premutation of the input corpus.}
  \label{fig:r-lex}
\end{center}
\end{figure}

The plots given in Figures~\ref{fig:pre}--\ref{fig:lex} are over
blocks of 500 utterances as discussed earlier.  However, because they
are a result of running the algorithm on a single corpus as
\shortciteA{Brent:EPS99} did, there is no way of telling if the
performance of each algorithm was influenced by any particular
ordering of the utterances in the corpus.  The question of whether the
algorithm is unduly biased by ordering idiosyncracies in the input
utterances was, in fact, also raised by one of the author's
colleagues.  A further undesirable effect of reporting results of a
run on exactly one ordering of the input is that there tends to be too
much variation between the values reported for consecutive scoring
blocks.  To account for both of these problems, we report averaged
results from running the algorithms on 1000 random permutations of the
input data.  This has the beneficial side-effect of allowing us to
plot with higher granularity since there is much less variation in the
precision and recall scores.  They are now clustered much closer to
their mean values in each block, allowing a block size of 100 to be
used to score the output.  These plots, given in
Figures~\ref{fig:r-pre}--\ref{fig:r-lex}, are much more readable than
those obtained before such averaging of the results.  

One may object that the original transcripts carefully preserve the
order of utterances directed at children by their mothers and hence
randomly permuting the corpus would destroy the fidelity of the
simulation.  However, as we argued, the permutation and averaging does
have significant beneficial side-effects, and if anything, it only
eliminates from the point of view of the algorithms the important
advantage that real children may be given by their mothers through a
specific ordering of the utterances.  In any case, we have found no
significant difference in performance between the permuted and
unpermuted cases as far as the various algorithms were concerned.


\subsection{Discussion}

Clearly, the performance of the present model is competitive with
MBDP-1 and as a consequence with other algorithms evaluated in
\shortciteA{Brent:EPS99}.  However, we note that the model proposed in
this paper has been entirely developed along conventional lines and
has not made the somewhat radical assumption of treating the entire
observed corpus as a single event in probability space.  Assuming that
the corpus consists of a single event, as Brent does, requires the
explicit calculation of the probability of the lexicon in order to
calculate the probability of any single segmentation.  This
calculation is a non-trivial task since one has to sum over all
possible orders of words in $\Lex$.  This fact is recognized in Brent,
where the expression for $\p(\Lex)$ is derived in Appendix~1 of his
paper as an approximation.  One can imagine then that it will be
correspondingly more difficult to extend the language model in
\shortciteA{Brent:EPS99} past the case of unigrams.  As a practical
issue, recalculating lexicon probabilities before each segmentation
also increases the running time of an implementation of the algorithm.
Although all the discussed algorithms tend to complete within a minute
on the corpus reported on, MBDP-1's running time is quadratic in the
number of utterances, while the language models presented here enable
computation in almost linear time.  The typical running time of MBDP-1
on the 9790 utterance corpus averages around 40 seconds per run on an
i686 PC while the 1-gram, 2-gram and 3-gram models average around 7,
10 and 14 seconds respectively.

Furthermore, the language models presented in this paper estimate
probabilities as relative frequencies using commonly used back-off
procedures and so they do not assume any priors over integers.
However, MBDP-1 requires the assumption of two distributions over
integers, one to pick a number for the size of the lexicon and another
to pick a frequency for each word in the lexicon.  Each is assumed
such that the probability of a given integer $\p(i)$ is given by
$\frac{6}{\pi^2n^2}$.  We have since found some evidence that suggests
that the choice of a particular prior does not have any significant
advantage over the choice of any other prior.  For example, we have
tried running MBDP-1 using $\p(i)=2^{-i}$ and still obtained
comparable results.  It is noteworthy, however, that no such
subjective prior needs to be chosen in the model presented in this
paper.

The other important difference between MBDP-1 and the present model is
that MBDP-1 assumes a uniform distribution over all possible word
orders.  That is, in a corpus that contains $n_k$ unique words such
that the frequency in the corpus of the $i$th unique word is given by
$f_k(i)$, the probability of any one ordering of the words in the
corpus is
$$
\frac{\prod_{i=1}^{n_k} f_k(i)!}{k!}
$$ 
because the number of unique orderings is precisely the reciprocal of
the above quantity.  Brent mentions that there may well be efficient
ways of using $n$-gram distributions in the same model.  The framework
presented in this paper is a formal statement of a model that lends
itself to such easy $n$-gram extensibility using the back-off scheme
proposed.  In fact, the results we present are direct extensions of
the unigram model into bigrams and trigrams.

\subsection{Responsiveness to training}

\begin{figure}[htb]
\begin{center}
  \includegraphics[width=11.8cm]{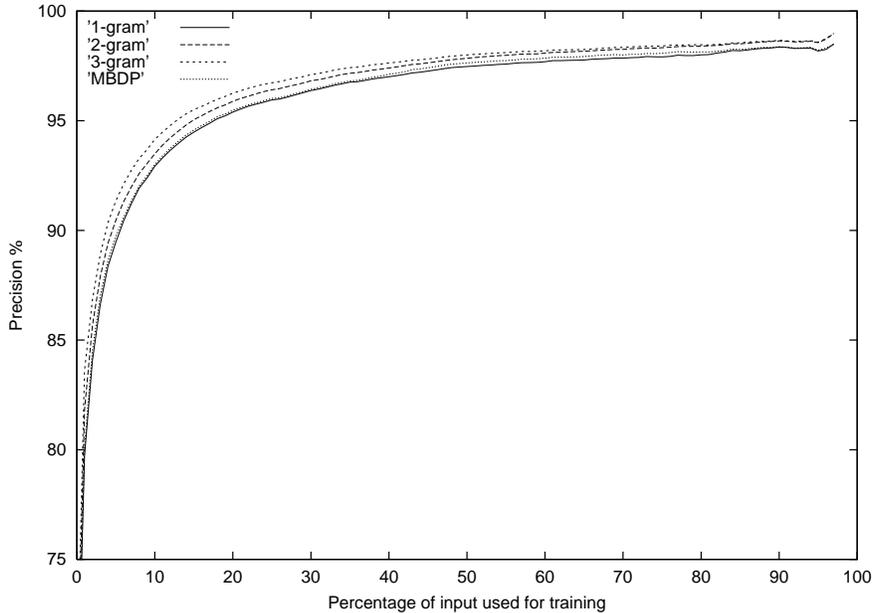}
  \caption{Responsiveness of the algorithm to training information.
  The horizontal axis represents the initial percentage of the data
  corpus that was used for training the algorithm.  This graph shows
  the improvement in precision with training size.}
  \label{fig:t-pre}
\end{center}
\end{figure}

\begin{figure}[htb]
\begin{center}
  \includegraphics[width=11.8cm]{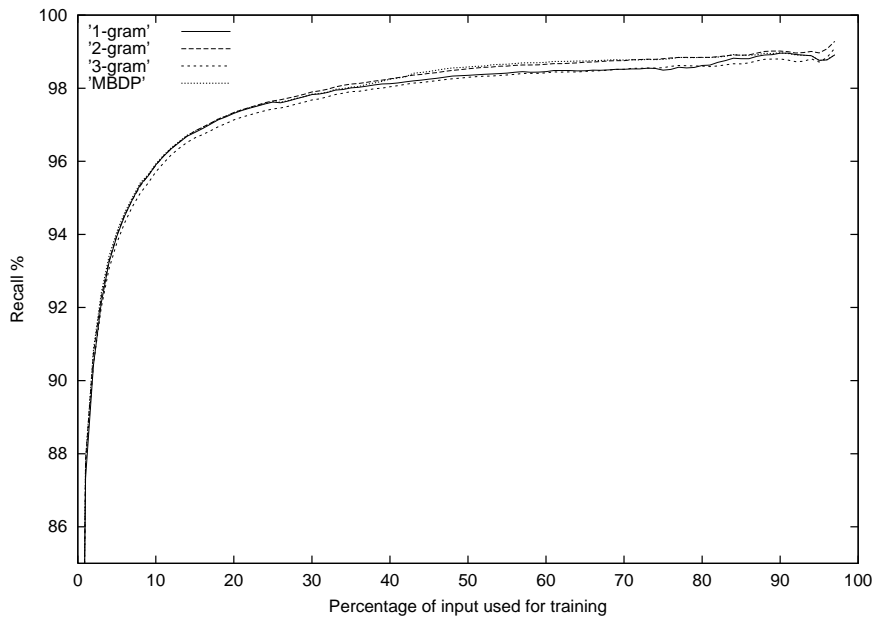}
  \caption{Improvement in recall with training size.}
  \label{fig:t-rec}
\end{center}
\end{figure}

\begin{figure}[htb]
\begin{center}
  \includegraphics[width=11.8cm]{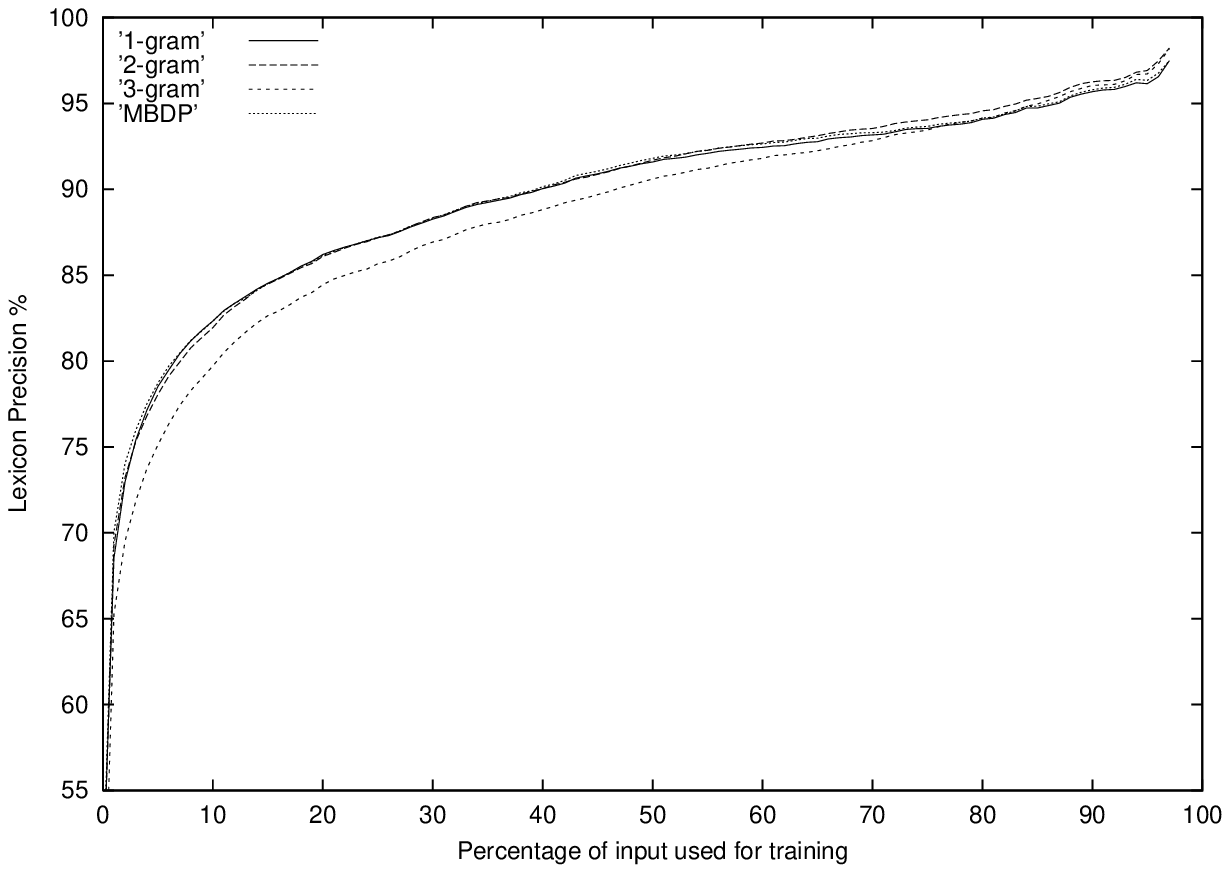}
  \caption{Improvement in lexicon precision with training size.}
  \label{fig:t-lex}
\end{center}
\end{figure}

It is interesting to compare the responsiveness of the various
algorithms to the effect of training data.
Figures~\ref{fig:t-pre}--\ref{fig:t-lex} plot the results (precision,
recall and lexicon precision) over the whole input corpus,
i.e. blocksize = $\infty$, as a function of the initial proportion of
the corpus reserved for training.  This is done by dividing the corpus
into two segments, with an initial training segment being used by the
algorithm to learn word, bigram, trigram and phoneme probabilities and
the latter actually being used as the test data.  A consequence of
this is that the amount of data available for testing becomes
progressively smaller as the percentage reserved for training grows.
So the significance of the test would diminish correspondingly.  We
may assume that the plots cease to be meaningful and interpretable
when more than about 75\% (about 7500 utterances) of the corpus is
used for training.  At 0 percent, there is no training information for
any algorithm and the scores are identical to those reported earlier.
We increase the amount of training data in steps of approximately 1
percent (100 utterances).  For each training set size, the results
reported are averaged over 25 runs of the experiment, each over a
separate random permutation of the corpus.  The motivation, as before,
was both to account for ordering idiosyncracies as well as to smooth
the graphs to make them easier to read.


We interpret Figures~\ref{fig:t-pre}--\ref{fig:t-lex} as suggesting
that increased history size contributes to increased precision.
Consequently, the 3-gram model is the most responsive to training
under this benchmark.  However, the scores obtained for recall and
lexicon precision require some explanation.  Clearly, while the
1-gram, 2-gram and MBDP-1 algorithms perform more or less similarly,
the 3-gram is seen to lag behind.  We suspect that this is due to the
peculiar nature of the domain in which there is a relatively larger
proportion of single word utterances.  Since the 3-gram model places
greatest emphasis on word triples, it has the least evidence of all
from the observed data to infer word boundaries.  Consequently, the
3-gram model is the most conservative in its predictions.  This is
consistent with the fact that its precision is high, whereas its
recall and lexicon scores are comparatively lower than the rest.  That
is, when it does have enough evidence to infer words, it places
boundaries in the right places, contributing to a high precision, but
more often than not, it simply does not output any segmentation, thus
outputting a single novel word (the entire utterance) instead of more
than one incorrectly inferred ones from it.  This contributes to its
poorer recall since recall is an indicator of the number of words the
model fails to infer.  Poorer lexicon precision is likewise explained.
Because the 3-gram model is more conservative, it only infers new
words when there is strong evidence for them.  As a result many
utterances are inserted as whole words into its lexicon thereby
contributing to decreased lexicon precision.  We further note that the
difference in performance between the different models tends to narrow
with increasing training size, i.e. as the amount of evidence
available to infer word boundaries increases, the 3-gram model rapidly
catches up with the others in recall and lexicon precision.  It is
likely, therefore, that with adequate training data, the 3-gram model
might be the most suitable one to use.  The following experiment lends
some substance to this suspicion.

\subsection{{\em Fully trained}\/ algorithms}

The preceding discussion makes us curious to see what would happen if
the above scenario was extended to the limit, i.e. if 100\% of the
corpus was used for training.  This precise situation was in fact
tested.  The entire corpus was concatenated onto itself and the models
then trained on exactly the former half and tested on the latter half
of the corpus augmented thus.  Although the unorthodox nature of this
procedure requires us to not attach much significance to the outcome,
we nevertheless find the results interesting enough to warrant some
mention and discussion here.  The performance of each of the four
algorithms on the test segment of the input corpus (the latter half)
is discussed below.  As one would expect from the results of the
preceding experiments, the trigram language model outperforms all
others.  It has a precision and recall of 100\% on the test input,
except for exactly four utterances.  These four utterances are shown
in Table~\ref{tbl:3gram-out}.
\begin{table}
\begin{center}
\begin{tabular}{|l|l|l|}
\hline
{\bf \#} & {\bf 3-gram output} & {\bf Target} \\
\hline \hline
3482 & $\cdots$ In D6 {\bf dOghQs} & $\cdots$ In D6 dOg hQs\\
5572 & {\bf 6klak}                             & 6 klak\\
5836 & D\&ts {\bf Olr9t}                       & D\&ts Ol r9t\\
7602 & D\&ts r9t Its 6 {\bf h*brAS}            & D\&ts r9t Its 6 h* brAS\\
\hline
\end{tabular}
\caption{Errors in the output of a fully trained 3-gram language
model.  Erroneous segmentations are shown in boldface.}
\label{tbl:3gram-out}
\end{center}
\end{table}

Intrigued as to why these errors occurred, we examined the corpus,
only to find erroneous transcriptions in the input.  ``Dog house'' is
transcribed as a single word ``dOghQs'' in utterance 614, and as two
words elsewhere.  Likewise, ``o'clock'' is transcribed ``6klAk'' in
utterance 5917, ``alright'' is transcribed ``Olr9t'' in utterance 3937
and ``hair brush'' is transcribed ``h*brAS'' in utterances 4838 and
7037.  Elsewhere in the corpus, these are transcribed as two words.

\begin{table}
\begin{center}
\begin{tabular}{|l|l|l|}
\hline
{\bf \#} & {\bf 2-gram output} & {\bf Target} \\
\hline \hline
 614 & yu want D6 {\bf dOg hQs}      & yu want D6 dOghQs\\
3937 & D\&ts {\bf Ol r9t}            & D\&ts Olr9t\\
5572 & {\bf 6klak}                   & 6 klak\\
7327 & lUk 6 {\bf h*brAS}            & lUk 6 h* brAS\\
7602 & D\&ts r9t Its 6 {\bf h*brAS}  & D\&ts r9t Its 6 h* brAS\\
7681 & {\bf h*brAS}                  & h* brAS\\
7849 & Its kOld 6 {\bf h*brAS}       & Its kOld 6 h* brAS\\
7853 & {\bf h*brAS}                  & h* brAS\\
\hline
\end{tabular}
\caption{Errors in the output of a fully trained 2-gram language
model.  The errorneous segmentations are shown in boldface.}
\label{tbl:2gram-out}
\end{center}
\end{table}

The erroneous segmentations in the output of the 2-gram language model
are also likewise shown in Table~\ref{tbl:2gram-out}.  As expected,
the effect of reduced history is apparent through an increase in the
total number of errors.  However, it is interesting to note that while
the 3-gram model incorrectly segmented an incorrect transcription
(utterance 5836) ``D\&ts Ol r9t'' to produce ``D\&ts Olr9t'', the
2-gram model incorrectly segmented a correct transcription (utterance
3937) ``D\&ts Olr9t'' to produce ``D\&ts Ol r9t''.  The reason for
this is that the bigram ``D\&ts Ol'' is encountered relatively
frequently in the corpus and this biases the algorithm towards
segmenting the ``Ol'' out of ``Olr9t'' when it follows ``D\&ts''.
However, the 3-gram model is not likewise biased because having
encountered the exact 3-gram ``D\&ts Ol r9t'' earlier, there is no
back-off to try bigrams at this stage.

Similarly, it is also interesting that while the 3-gram model
incorrectly segments the incorrectly transcribed ``dOg hQs'' into
``dOghQs'' in utterance 3482, the 2-gram model incorrectly segments
the correctly transcribed ``dOghQs'' into ``dOg hQs'' in utterance
614.  In the trigram model, $-\log \p({\rm hQs|D6,dOg}) = 4.77569$ and
$-\log \p({\rm dOg|In, D6}) = 5.3815$, giving a score of 10.1572 to the
segmentation ``dOg hQs''.  However, due to the error in transcription,
the trigram ``In D6 dOghQs'' is never encountered in the training data
although the bigram ``D6 dOghQs'' is.  Backing off to bigrams, $-\log
\p({\rm dOghQs|D6})$ is calculated as 8.12264.  Hence the probability
that ``dOghQs'' is segmented as ``dOg hQs'' is less than the
probability that it is a word by itself.  In the 2-gram model, $-\log
\p({\rm dOg|D6}) -\log \p({\rm hQs|dOg}) = 3.67979 + 3.24149 = 6.92128$
whereas $-\log \p({\rm dOghQs|D6}) = 7.46397$, whence ``dOghQs'' is the
preferred segmentation although the training data contained instances
of all three bigrams.

\begin{table}
\begin{center}
\begin{tabular}{|l|l|l|}
\hline
{\bf \#} & {\bf 1-gram output} & {\bf Target} \\
\hline \hline
 244 & brAS {\bf \&lIs Iz} h*                 & brAS \&lIsIz h*\\
 503 & y) {\bf In tu} dIstrAkS$\sim$ $\cdots$      & y) Intu dIstrAkS$\sim$ $\cdots$\\
1066 & yu {\bf m9 trIp} It                    & yu m9t rIp It\\
1231 & DIs Iz lItL {\bf dOghQs}               & DIs Iz lItL dOg hQs\\
1792 & stIk It {\bf an tu} D*                 & stIk It antu D*\\
3056 & $\cdots$ so hi dAz$\sim$t rAn {\bf In tu}   & $\cdots$ so hi dAz$\sim$t rAn Intu\\
3094 & $\cdots$ tu bi In D6 {\bf h9c*}        & $\cdots$ tu bi In D6 h9 c*\\
3098 & $\cdots$ f\% DIs {\bf h9c*}            & $\cdots$ f\% DIs h9 c*\\
3125 & $\cdots$ {\bf OlrEdi} $\cdots$         & $\cdots$ Ol rEdi $\cdots$ \\
3212 & $\cdots$ k6d tOk {\bf In tu} It        & $\cdots$ k6d tOk Intu It\\
3230 & k\&n {\bf hil 9} dQn an DEm            & k\&n hi l9 dQn an DEm\\
3476 & D\&ts 6 {\bf dOghQs}                   & D\&ts 6 dOg hQs	\\
3482 & $\cdots$ In D6 {\bf dOghQs}            & $\cdots$ In D6 dOg hQs\\
3923 & $\cdots$ WEn {\bf Its noz}             & $\cdots$ WEn It snoz\\
3937 & D\&ts {\bf Ol r9t}                     & D\&ts Olr9t\\
4484 & Its 6bQt {\bf milt9m z}                & Its 6bQt mil t9mz\\
5328 & tEl hIm tu {\bf we kAp}                & tEl hIm tu wek Ap\\
5572 & {\bf 6klak}                            & 6 klak\\
5671 & W*z m9 lItL {\bf h*brAS}               & W*z m9 lItL h* brAS\\
6315 & D\&ts {\bf 6 ni}                       & D\&ts 6n i\\
6968 & oke mami {\bf tek sIt}                 & oke mami teks It\\
7327 & lUk 6 {\bf h*brAS}                     & lUk 6 h* brAS\\
7602 & D\&ts r9t Its 6 {\bf h*brAS}           & D\&ts r9t Its 6 h* brAS\\
7607 & go {\bf 6lON} we tu f9nd It t6de       & go 6 lON we tu f9nd It t6de\\
7676 & mam {\bf pUt sIt}                      & mam pUts It\\
7681 & {\bf h*brAS}                           & h* brAS\\
7849 & Its kOld 6 {\bf h*brAS}                & Its kOld 6 h* brAS\\
7853 & {\bf h*brAS}                           & h* brAS\\
8990 & $\cdots$ In D6 {\bf h9c*}              & $\cdots$ In D6 h9 c*\\
8994 & f\% bebiz 6 n9s {\bf h9c*}             & f\% bebiz 6 n9s h9 c*\\
8995 & D\&ts l9k 6 {\bf h9c*} D\&ts r9t       & D\&ts l9k 6 h9 c* D\&ts r9t\\
9168 & hi h\&z {\bf 6lON} tAN                 & hi h\&z 6 lON tAN\\
9567 & yu wan6 go In D6 {\bf h9c*}            & yu wan6 go In D6 h9 c*\\
9594 & {\bf 6lON} rEd tAN                     & 6 lON rEd tAN\\
9674 & {\bf dOghQs}                           & dOg hQs\\
9688 & {\bf h9c*} 6gEn                        & h9 c* 6gEn\\
9689 & $\cdots$ D6 {\bf h9c*}                 & $\cdots$ D6 h9 c*\\
9708 & 9 h\&v {\bf 6lON} tAN                  & 9 h\&v 6 lON tAN\\
\hline
\end{tabular}
\caption{Errors in the output of a fully trained 1-gram language model.}
\label{tbl:1gram-out}
\end{center}
\end{table}

The errors in the output of a 1-gram model are also shown in
Table~\ref{tbl:1gram-out}, but they are not discussed as we did above
for the 3-gram and 2-gram outputs.  The errors in the output of
Brent's fully-trained MBDP-1 algorithm are not shown here because they
are identical to those produced by the 1-gram model except for one
utterance.  This only difference is the segmentation of utterance
8999, ``lItL QtlEts'' (little outlets), which the 1-gram model
segmented correctly as ``lItL QtlEts'', but MBDP-1 segmented as ``lItL
{\bf Qt lEts}''.  In both MBDP-1 and the 1-gram model, all four words,
``little'', ``out'', ``lets'' and ``outlets'' are familiar at the time
of segmenting this utterance.  MBDP-1 assigns a score of $5.29669 +
5.95011 = 11.2468$ to the segmentation ``out + lets'' versus a score
of $11.7613$ to ``outlets''.  As a consequence, ``out + lets'' is the
preferred segmentation.  In the 1-gram language model, the
segmentation ``out + lets'' scores $5.31399 +  5.96457 = 11.27856$,
whereas ``outlets'' scores $11.0885$.  Consequently it selects
``outlets'' as the preferred segmentation.  The only thing we could
surmise from this was either that this difference must have come about
due to chance (meaning that this may well have not been the case if
certain parts of the corpus had been any different) or else the
interplay between the different elements in the two models is too
subtle to be addressed within the scope of this paper.

\subsection{Similarities between MBDP-1 and the 1-gram Model}

The similarities between the output of MBDP-1 and the 1-gram model are
so great as to suspect that they may essentially be capturing the same
nuances of the domain.  Although \shortciteA{Brent:EPS99} explicitly
states that probabilities are not estimated for words, it turns out
that considering the entire corpus does end up having the same effect
as estimating probabilities from relative frequencies as the 1-gram
model does.  The {\em relative probability}\/ of a familiar word is
given in Equation~22 of \shortciteA{Brent:EPS99} as
$$
\frac{f_k(\hat{k})}{k}\cdot
\left( \frac{f_k(\hat{k})-1}{f_k(\hat{k})}\right)^2
$$
where $k$ is the total number of words and $f_k(\hat{k})$ is the
frequency at that point in segmentation of the $k$th word. It 
effectively approximates to the relative frequency 
$$
\frac{f_k(\hat{k})}{k}
$$
as $f_k(\hat{k})$ grows.  The 1-gram language model of this paper
explicitly claims to use this specific estimator for the unigram
probabilities.  From this perspective, both MBDP-1 and the 1-gram
model tend to favor the segmenting out of familiar words that do not
overlap.  It is interesting, however, to see exactly how much evidence
each needs before such segmentation is carried out.  In this context,
the author recalls an anecdote recounted by a British colleague who
while visiting the USA, noted that the populace in the vicinity of his
institution grew up thinking that ``Damn British'' was a single word,
by virtue of the fact that they had never heard the latter word in
isolation.  We test this particular scenario here with both
algorithms.  The program is first presented with the utterance ``D\&m
brItIS''.  Having no evidence to infer otherwise, both programs assume
that ``D\&mbrItIS'' is a single word and update their lexicons
accordingly.  The interesting question now is exactly how many
instances of the word ``British'' in isolation should either program
see before being able to successfully segment a subsequent
presentation of ``Damn British'' correctly.

Obviously, if the word ``D\&m'' is also unfamiliar, there will never
be enough evidence to segment it out in favor of the familiar word
``D\&mbrItIS''.  Hence each program is presented next with two
identical utterances, ``D\&m''.  We do need to present two such
utterances.  Otherwise the estimated probabilities of the familiar
words ``D\&m'' and ``D\&mbrItIS'' will be equal.  Consequently, the
probability of any segmentation of ``D\&mbrItIS'' that contains the
word ``D\&m'' will be less than the probability of ``D\&mbrItIS''
considered as a single word.

At this stage, we present each program with increasing numbers of
utterances consisting solely of the word ``brItIS'' followed by a
repetition of the very first utterance -- ``D\&mbrItIS''.  We find
that MBDP-1 needs to see the word ``brItIS'' on its own three times
before having enough evidence to disabsuse itself of the notion that
``D\&mbrItIS'' is a single word.  In comparison, the 1-gram model is
more skeptical.  It needs to see the word ``brItIS'' on its own seven
times before committing to the right segmentation.  It is easy to
predict this number analytically from the presented 1-gram model, for
let $x$ be the number of instances of ``brItIS'' required.  Then using
the discounting scheme described, we have 
\begin{eqnarray*}
\p({\rm D\&mbrItIS}) &=& 1/(x+6)\\
\p({\rm D\&m}) &=& 2/(x+6)\qquad {\rm  and}\\
\p({\rm brItIS}) &=& x/(x+6)
\end{eqnarray*}
We seek an $x$ for which $\p({\rm D\&m}) \p({\rm brItIS}) > \p({\rm
D\&mbrItIS})$.  Thus, we get
$$
2x/(x+6)^2 > 1/(x+6) \Rightarrow x > 6
$$
The actual scores for MBDP-1 when presented with ``D\&mbrItIS'' for a
second time are: $-\log \p({\rm D\&mbrItIS}) = 2.77259$ and $-\log
\p({\rm D\&m}) -\log \p({\rm brItIS}) = 1.79176 + 0.916291 = 2.70805$.
For the 1-gram model, $-\log \p({\rm D\&mbrItIS}) = 2.56495$ and $-\log
\p({\rm D\&m}) -\log \p({\rm brItIS}) = 1.8718 + 0.619039 = 2.49084$.
Note, however, that skepticism in this regard is not always a bad
attribute.  It helps to be skeptical in inferring new words because a
badly inferred word will adversely influence future segmentation
accuracy.

\section{Summary}
\label{sec:summary}

In summary, we have presented a formal model of word discovery in
child directed speech.  The main advantages of this model over those
of \shortciteA{Brent:EPS99} are firstly that the present model has
been developed entirely by direct application of standard techniques
and procedures in speech processing.  It also makes few assumptions
about the nature of the domain and remains as far as possible
conservative in its development.  Finally, the model is easily
extensible to incorporate more historical detail.  This is clearly
evidenced by the extension of the unigram model to handle bigrams and
trigrams.  Empirical results from experiments suggest that the model
performs competitively with alternative models currently in use for
the purpose of inferring words from fluent child-directed speech.

Although the algorithm is originally presented as an unsupervised
learner, we have also shown the effect that training data has on its
performance.  It appears that the 3-gram model is the most responsive
to training information with regard to segmentation precision,
obviously by virtue of the fact that it {\em keeps\/} more knowledge
from the presented utterances.  Indeed, we see that a fully-trained
3-gram model performs with 100 percent accuracy on the test set.
Admittedly, the test set in this case was identical to the training
set, but we should keep in mind that we were also only keeping limited
history, namely 3-grams, and a significant number of utterances in the
input corpus (4023 utterances) were 4 words or more in length.  Thus
it is not completely insignificant that the algorithm was able to
perform this well.

\subsection{Future work}

It is tempting to extend the approach presented in this paper to
handle domains other than child directed speech.  This is in part
constrained by the lack of availability of phonemically transcribed
speech in these other domains.  However, it has been suggested that we
should be able to test the performance of a trained algorithm on
speech data from say, the Switchboard telephone speech corpus.  Such
work is, in fact, in the process of being investigated at the present
time.

Further extensions being worked on include the incorporation of more
complex phoneme distributions into the model.  These are, namely, the
biphone and triphone models.  Some preliminary results we have
obtained in this regard appear to be encouraging.
\shortciteA[p.101]{Brent:EPS99} remarks that learning phoneme
probabilities from lexical entries yielded better results than
learning these probabilities from speech.  That is, the probability of
the phoneme ``th'' in ``the'' is better not inflated by the
preponderance of {\em the}\/ and {\em the}-like words in actual
speech, but rather controlled by the number of such unique words.  We
are unable to confirm this in the domain of child-directed speech with
either our analysis or our experiments.  For assume the existence of
some function $\Psi_X:{\bf N}\rightarrow{\bf N}$ that maps the size,
$n$, of a corpus $\C$, onto the size of some subset $\X$ of $\C$ we
may define.  If this subset $\X=\C$, then $\Psi_\C$ is the identity
function and if $\X=\Lex$ is the set of unique words in $\C$ we have
$\Psi_\Lex(n) = |\Lex|$.

\begin{figure}[htb]
\begin{center}
  \includegraphics[width=11.8cm]{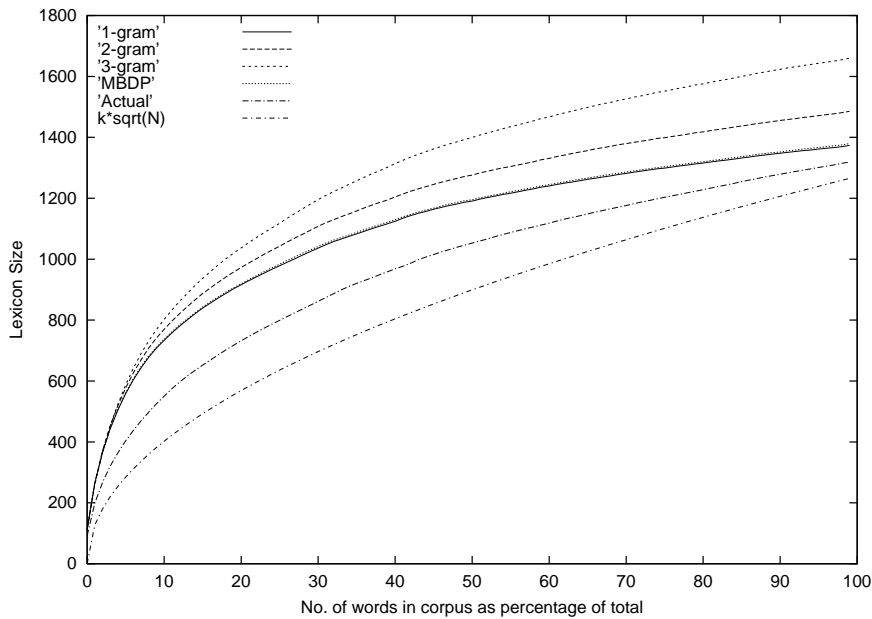} 
  \caption{Plot shows the rate of growth of the lexicon with
  increasing corpus size as percentage of total size.  {\em Actual}\/
  is the actual number of unique words in the input corpus.  {\em 1-gram,
  2-gram, 3-gram}\/ and {\em MBDP}\/ plot the size of the lexicon
  as inferred by each of the algorithms.  It is interesting that the
  rate of lexicon growth is roughly similar regardless of the algorithm
  used to infer words and that they may all potentially be modeled by
  a function such as $k\sqrt{N}$ where $N$ is the corpus size.}
  \label{fig:lexrate}
\end{center}
\end{figure}

Let $l_\X$ be the average number of phonemes per word in $\X$ and let
${E_a}_\X$ be the average number of occurrences of phoneme $a$ per
word in $\X$.  Then we may estimate the probability of an arbitrary
phoneme $a$ from $\X$ as:
\begin{eqnarray*}
\p(a|\X) &=& \frac{C(a|\X)}{\sum_{a_i} C(a_i|\X)} \\
         &=& \frac{{E_a}_\X \Psi_\X(N)}{l_\X\Psi_\X(N)}
\end{eqnarray*}
where, as before, $C(a|\X)$ is the count function that gives the
frequency of phoneme $a$ in $\X$.  If $\Psi_X$ is deterministic, we
can then write
\begin{equation}
\p(a|\X) = \frac{{E_a}_\X}{l_\X} \label{eqn:phon-Xprob}
\end{equation}
Our experiments in the domain of child directed speech suggest that
${E_a}_\Lex \sim {E_a}_\C$ and that $l_\Lex \sim l_\C$.  We are thus
led to suspect that estimates should roughly be the same regardless of
whether probabilities are estimated from $\Lex$ or $\C$.  This is
indeed borne out by the results we present below.  Of course, this is
only true if there exists some deterministic function $\Psi_\Lex$ as
we assumed and this may not necessarily be the case.  There is,
however, some evidence that the number of unique words in a corpus can
be related to the total number of words in the corpus in this way.  In
Figure~\ref{fig:lexrate} the rate of lexicon growth is plotted against
the proportion of the corpus size considered.  The values for lexicon
size were collected using the Unix filter
\begin{quote}
{\tt cat \$*|tr ' ' $\backslash\backslash$012|awk '}\{{\tt print
(L[\$0]++)? v : ++v;}\}'
\end{quote}
and smoothed by averaging over 100 runs each on a separate permutation
of the input corpus.  That the lexicon size can be approximated by a
deterministic function of the corpus size in the domain of child
directed speech is strongly suggested by the the plot.  In particular,
we suspect the function $\Psi$ to be of the form $k\sqrt{|\C|}$ for a
given corpus $\C$.  In this case, $k$ happens to be 7.
Interestingly, the shape of the plot is roughly the same regardless of
the algorithm used to infer words suggesting that they all segment
{\em word-like}\/ units which share at least some statistical
properties with actual words.

Table~\ref{tbl:phon-learn} summarizes our empirical findings in this
regard.  For each model, namely 1-gram, 2-gram, 3-gram and MBDP-1, we
test all three of the following possibilities:
\begin{enumerate}
\item Always use a uniform distribution over phonemes
\item Learn the phoneme distribution from the lexicon and
\item Learn the phoneme distribution from speech, i.e. from the words
in the corpus, whether unique or not.
\end{enumerate}

\begin{table}
\begin{center}
\begin{tabular}{l|l|l|l|l|}
\multicolumn{5}{c}{} \\
\multicolumn{1}{c}{}   & \multicolumn{4}{c}{\bf Precision} \\ 
                                                   \cline{2-5}
\multicolumn{1}{c|}{}  & 1-gram  & 2-gram  & 3-gram  & MBDP      \\ 
                                                   \cline{2-5}
{\bf Lexicon}          & 67.7  & 68.08 & 68.02 & 67        \\
{\bf Speech}           & 66.25 & 66.68 & 68.2  & 66.46     \\
{\bf Uniform}          & 58.08 & 64.38 & 65.64 & 57.15     \\
\cline{2-5}\cline{2-5}
\multicolumn{5}{c}{}\\
\multicolumn{1}{c}{}   & \multicolumn{4}{c}{\bf Recall }   \\ 
                                                   \cline{2-5}
\multicolumn{1}{c|}{}  & 1-gram  & 2-gram  & 3-gram  & MBDP      \\
                                                   \cline{2-5}
{\bf Lexicon}          & 70.18 & 68.56 & 65.07 & 69.39     \\
{\bf Speech}           & 69.33 & 68.02 & 66.06 & 69.5      \\
{\bf Uniform}          & 65.6  & 69.17 & 67.23 & 65.07     \\
\cline{2-5}\cline{2-5}
\multicolumn{5}{c}{}\\
\multicolumn{1}{c}{}   & \multicolumn{4}{c}{\bf Lexicon Precision}   \\
                                                   \cline{2-5}
\multicolumn{1}{c|}{}  & 1-gram  & 2-gram  & 3-gram  & MBDP      \\
                                                   \cline{2-5}
{\bf Lexicon}          & 52.85 & 54.45 & 47.32 & 53.56     \\
{\bf Speech}           & 52.1  & 54.96 & 49.64 & 52.36     \\
{\bf Uniform}          & 41.46 & 52.82 & 50.8  & 40.89     \\
\cline{2-5}\cline{2-5}
\end{tabular}
\caption{Summary of results from each of the algorithms for each of
the following cases:  Lexicon -- Phoneme probabilities estimated from
the lexicon, Speech -- Phoneme probabilities estimated from input
corpus and Uniform -- Phoneme probabilities are assumed uniform and
constant.}
\label{tbl:phon-learn}
\end{center}
\end{table}
The row labeled {\em Lexicon}\/ lists scores on the entire corpus from
a program that learned phoneme probabilities from the lexicon.  The
row labeled {\em Speech}\/ lists scores from a program that learned
these probabilities from speech, and the row labeled {\em Uniform}\/
lists scores from a program that just assumed uniform phoneme
probabilities throughout.

While the lexicon precision is clearly seen to suffer when a uniform
distribution over phonemes is assumed for MBDP-1, whether the
distribution is estimated from the lexicon or speech data does not
seem to make any significant difference.  Indeed, the recall is
actually seen to improve marginally if the phoneme distribution is
estimated from speech data as opposed to lexicon data.  These results
lead us to believe, contrary to the claim in \shortciteA{Brent:EPS99},
that it really doesn't matter whether phoneme probabilities are
estimated from the corpus or the lexicon.

Expectedly, the 1-gram model also behaves similarly to MBDP-1.
However, both the 2-gram and 3-gram models seem much more robust in
face of changing methods of estimating phoneme probabilities.  We
assumed initially that this was probably due to the fact that as the
size of the history increased, the number of back-offs one had to
perform in order to reach the level of phonemes was correspondingly
greater and so that much lesser should have been the significance of
assuming any particular distribution over phonemes.  But as the
results in Table~\ref{tbl:phon-learn} show, the evidence at this
point is too meager to give specific explanations for this.

With regard to estimation of word probabilities, modification of the
model to address the sparse data problem using interpolation such that
$$
\p(w_i|w_{i-2},w_{i-1}) = \lambda_3 \p(w_i|w_{i-2},w_{i-1}) + \lambda_2
	\p(w_i|w_{i-1}) + \lambda_1 \p(w_i)
$$
where the positive coefficients satisfy $\lambda_1 + \lambda_2 +
\lambda_3 = 1$ and can be derived so as to maximize $\p(\W)$ is also
being considered as a fruitful avenue.

Using the lead from \shortciteA{Brent:EPS99}, attempts to model more
complex distributions for unigrams such as those based on {\em
template grammars\/} or the incorporation of prosodic, stress and
phonotactic constraint information into the model are also the subject
of current interest.  We already have some unpublished results which
suggest that biasing the segmentation towards segmenting out words
which conform to given templates (such as CVC for {\em Consonant,
Vowel, Consonant}) greatly increases segmentation accuracy.  In fact,
imposing a constraint that every word must have at least one vowel in
it dramatically increases segmentation precision from 67.7\% to 81.8\%
and imposing a constraint that words can only begin or end with
permitted clusters of consonants increases precision to 80.65\%.
Experiments are underway to investigate models in which these
templates can be learned in the same way as $n$-grams.  Finally, work
on incorporating an acoustic model into the picture so as to be able
to calculate $\p(\A|\W)$ is also being looked at.  Since $\p(\A|\W)$
and $\p(\W)$ are generally believed to be independent, work in each
component can proceed more or less in parallel.

\section*{Acknowledgments}

The author wishes to thank {\bf Michael Brent} for initially
introducing him to the problem, for several stimulating discussions on
the topic and many valuable suggestions.  Thanks are also due to {\bf
Koryn Grant} for cross-checking the results presented here and
suggesting some extensions to the model that are presently being
worked on.  The anecdote about the {\em Damn British}\/ is due to
{\bf Robert Linggard}. {\bf Claire Cardie} contributed significantly
by way of constructive criticism of a very early version of this
paper.

\bibliography{mml,seg}
\bibliographystyle{theapa}

\newpage
\section*{Appendix A - Inventory of Phonemes}

The following tables list the ASCII representations of the phonemes
used to transcribe the corpus into a form suitable for processing by
the algorithms.

\begin{singlespace}

\begin{tabular}{|c|l|} 
\multicolumn{2}{c}{\bf Consonants}\\ \hline
{\bf ASCII} & {\bf Example}  \\ \hline \hline
p  & {\bf p}an \\
b  & {\bf b}an \\
m  & {\bf m}an \\
t  & {\bf t}an \\
d  & {\bf d}am \\
n  & {\bf n}ap \\
k  & {\bf c}an \\
g  & {\bf g}o \\
N  & si{\bf ng} \\
f  & {\bf f}an \\
v  & {\bf v}an \\
T  & {\bf th}in \\
D  & {\bf th}an \\
s  & {\bf s}and \\
z  & {\bf z}ap \\
S  & {\bf sh}ip \\
Z  & plea{\bf s}ure \\
h  & {\bf h}at \\
c  & {\bf ch}ip \\
G  & {\bf g}el \\
l  & {\bf l}ap \\
r  & {\bf r}ap \\
y  & {\bf y}et \\
W  & {\bf wh}en \\
L  & bott{\bf le} \\
M  & rhyth{\bf m} \\
$\sim$  &  butt{\bf on} \\ \hline
\end{tabular}
\begin{tabular}{|c|l|} 
\multicolumn{2}{c}{\bf Vowels}\\ \hline
{\bf ASCII} & {\bf Example}  \\ \hline \hline
I  & b{\bf i}t \\
E  & b{\bf  e}t \\
\& & b{\bf  a}t \\
A  & b{\bf  u}t \\
a  & h{\bf  o}t \\
O  & l{\bf  aw} \\
U  & p{\bf  u}t \\
6  & h{\bf  e}r \\
i  & b{\bf ee}t \\
e  & b{\bf ai}t \\
u  & b{\bf oo}t\\
o  & b{\bf oa}t \\
9  & b{\bf uy} \\
Q  & b{\bf ou}t \\
7  & b{\bf oy} \\\hline
\end{tabular}
\begin{tabular}{|c|l|}
\multicolumn{2}{c}{\bf Vowel + r}\\ \hline
{\bf ASCII} & {\bf Example}  \\ \hline \hline
3  & b{\bf ir}d\\
R  & butt{\bf er} \\
\# & {\bf ar}m \\
\% & h{\bf or}n \\
*  & {\bf air} \\
(  & {\bf ear} \\
)  & l{\bf ure} \\ \hline
\end{tabular}

\end{singlespace}

\end{document}